\documentclass[10pt,twocolumn,letterpaper]{article}

\usepackage{cvpr}              % To produce the CAMERA-READY version
\definecolor{cvprblue}{rgb}{0.21,0.49,0.74}
\usepackage{multirow}
\usepackage{rotating}
\usepackage{booktabs}

\def\paperID{12726} % *** Enter the Paper ID here
\def\confName{CVPR}
\def\confYear{2025}

\title{Stochastic Human Motion Prediction with Memory of Action \\ Transition and Action Characteristic}

\author{
Jianwei Tang$^{1\star}$ \quad 
Hong Yang$^{1\star}$ \quad 
Tengyue Chen$^{1}$ \quad
Jian-Fang Hu$^{1,2,3\dagger}$ \\
$^{1}$School of Computer Science and Engineering, Sun Yat-sen University, China \\
$^{2}$Guangdong Province Key Laboratory of Information Security Technology, China \\
$^{3}$Key Laboratory of Machine Intelligence and Advanced Computing, Ministry of Education, China \\
{\tt\small \{tangjw7,yangh553,chenty63\}@mail2.sysu.edu.cn, hujf5@mail.sysu.edu.cn}
}

\begin{document}
\maketitle
\let\thefootnote\relax\footnotetext{$\star$ Equal contribution.}
\let\thefootnote\relax\footnotetext{$\dagger$ Corresponding author.}
% \begin{teaserfigure}
%   \centering
%    \includegraphics[width=0.95\linewidth]{fig/intro.pdf}
%    \caption{TEMP}
%    \label{fig:intro}
% \end{teaserfigure}

\begin{abstract}
Action-driven stochastic human motion prediction aims to generate future motion sequences of a pre-defined target action based on given past observed sequences performing non-target actions. This task primarily presents two challenges. Firstly, generating smooth transition motions is hard due to the varying transition speeds of different actions. Secondly, the action characteristic is difficult to be learned because of the similarity of some actions. These issues cause the predicted results to be unreasonable and inconsistent. As a result, we propose two memory banks, the Soft-transition Action Bank (STAB) and Action Characteristic Bank (ACB), to tackle the problems above. The STAB stores the action transition information. It is equipped with the novel soft searching approach, which encourages the model to focus on multiple possible action categories of observed motions. The ACB records action characteristic, which produces more prior information for predicting certain actions. To fuse the features retrieved from the two banks better, we further propose the Adaptive Attention Adjustment (AAA) strategy. Extensive experiments on four motion prediction datasets demonstrate that our approach consistently outperforms the previous state-of-the-art. The demo and code are available at \url{https://hyqlat.github.io/STABACB.github.io/}.
\end{abstract}
% It further prevents model to learn a correct transition from observed frames to target action.    
\section{Introduction}
\label{sec:intro}
Action-driven stochastic human motion prediction \cite{mao2022weakly} aims to predict future motion sequences based on past observed sequences with certain future action labels. This task is widely used in many applications like virtual reality \cite{koppula2013anticipating} and human-computer interaction \cite{starke2019neural}. Solving this problem can play a significant role in enabling artificial intelligence to make predictions and responses in advance. However, generating smooth transition motions is difficult due to the varying transition speeds of different actions and the action characteristic is hard to be learned. As a result, the predicted results may be inconsistent and unreasonable. 
\begin{figure}[t]
  \centering
  % \fbox{\rule{0pt}{2in} \rule{0.95\linewidth}{0pt}}
   \includegraphics[width=0.88\linewidth]{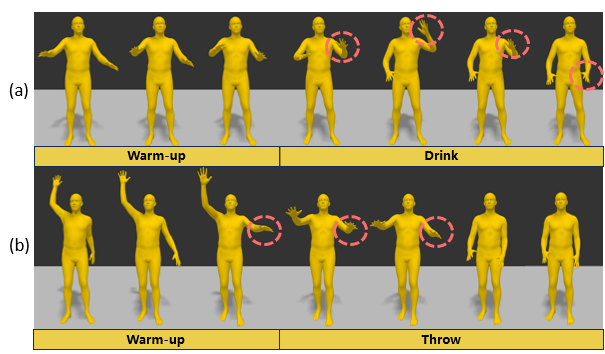}
   \vspace{-1.2mm}
   \caption{
Limitations of WAT \cite{mao2022weakly} in predicting action-driven stochastic human motion. (a) shows the transition from ``Warm-up'' to ``Drink'', where the generated ``Drink'' action is inaccurate, resembling a hand-raising motion. (b) shows the transition from ``Warm-up'' to ``Throw'', where the left hand moves naturally, but the arm fails to lift, resulting in uncoordinated movement.}
    \vspace{-3.2mm}
   \label{fig:intro}
\end{figure}

Some researchers are dedicated to addressing these issues. Ye \etal \cite{yuan2020dlow} design a loss function that encourages the model to generate more diverse motion sequences. Wei \etal \cite{mao2022weakly} proposed to predict future motion controlled by the action labels. However, these models still have some shortcomings. They did not make good use of the action transition information and the action characteristics, making it difficult for the model to generate smooth motion sequences and pay attention to these subtle differences.

Take the samples in Figure \ref{fig:intro} as an example, in which (a) illustrates the generation from the ``Warm-up'' action to the ``Drink'' action. However, the generated drink action is somewhat inaccurate. It resembles a hand-raising motion. This may be due to the high similarity between the two actions, making it challenging for prior methods to achieve finer differentiation between them. (b) illustrates the generation from the ``Warm-up'' action to the ``Throw'' action. In the ``Throw'' action, the left hand transitions naturally, but the arm fails to lift accordingly, lacking coordinated movement. This may result from insufficient consideration of natural transitions between actions. These aspects are crucial in applications such as human-computer interaction \cite{starke2019neural}, animation \cite{van2010real}, and motion planning \cite{la2011motion}, which rely on seamless and precise action transitions to enable realistic and accurate task execution.

To tackle the problems mentioned above, we design two memory banks to store action transition information and action characteristics, the soft-transition action bank (STAB) and action characteristic bank (ACB). The STAB is proposed to record action transition information of different actions. We equip its querying process with a soft searching approach. Since some actions have similar parts, there are multiple candidate possible action categories in the observed sequence. Therefore, the proposed soft searching method can encourage the model to focus on multiple possibilities of observed action. We further propose the ACB to preserve action characteristics for retrieving, which provides more detailed information for the generation of long sequences. To fuse features retrieved from two banks better, we propose the Adaptive Attention Adjustment (AAA) strategy. This strategy promotes the model to utilize more necessary feature information at different times, \ie action transition features for the initial period of prediction while action characteristics for the later prediction period.

Our contribution can therefore be summarized as follows: (i) We propose two banks STAB and ACB to record action transition features and action characteristics, which promotes the prediction performance of the baseline model. (ii) We design a novel feature fusion approach AAA that effectively utilizes action information at each predicting step. (iii) The sufficient experimental results have verified the effectiveness of our method and achieved SOTA results on four motion prediction datasets.
% \begin{figure}[t]
%   \centering
%   \fbox{\rule{0pt}{2in} \rule{0.9\linewidth}{0pt}}
%    %\includegraphics[width=0.8\linewidth]{egfigure.eps}
%    \caption{Example of caption.
%    It is set in Roman so that mathematics (always set in Roman: $B \sin A = A \sin B$) may be included without an ugly clash.}
%    \label{fig:onecol}
% \end{figure}
% \begin{figure*}
%   \centering
%   \begin{subfigure}{0.68\linewidth}
%     \fbox{\rule{0pt}{2in} \rule{.9\linewidth}{0pt}}
%     \caption{An example of a subfigure.}
%     \label{fig:short-a}
%   \end{subfigure}
%   \hfill
%   \begin{subfigure}{0.28\linewidth}
%     \fbox{\rule{0pt}{2in} \rule{.9\linewidth}{0pt}}
%     \caption{Another example of a subfigure.}
%     \label{fig:short-b}
%   \end{subfigure}
%   \caption{Example of a short caption, which should be centered.}
%   \label{fig:short}
% \end{figure*}
\begin{figure*}[htbp]
  \centering
  % \fbox{\rule{0pt}{2in} \rule{0.95\linewidth}{0pt}}
   \includegraphics[width=0.8\linewidth]{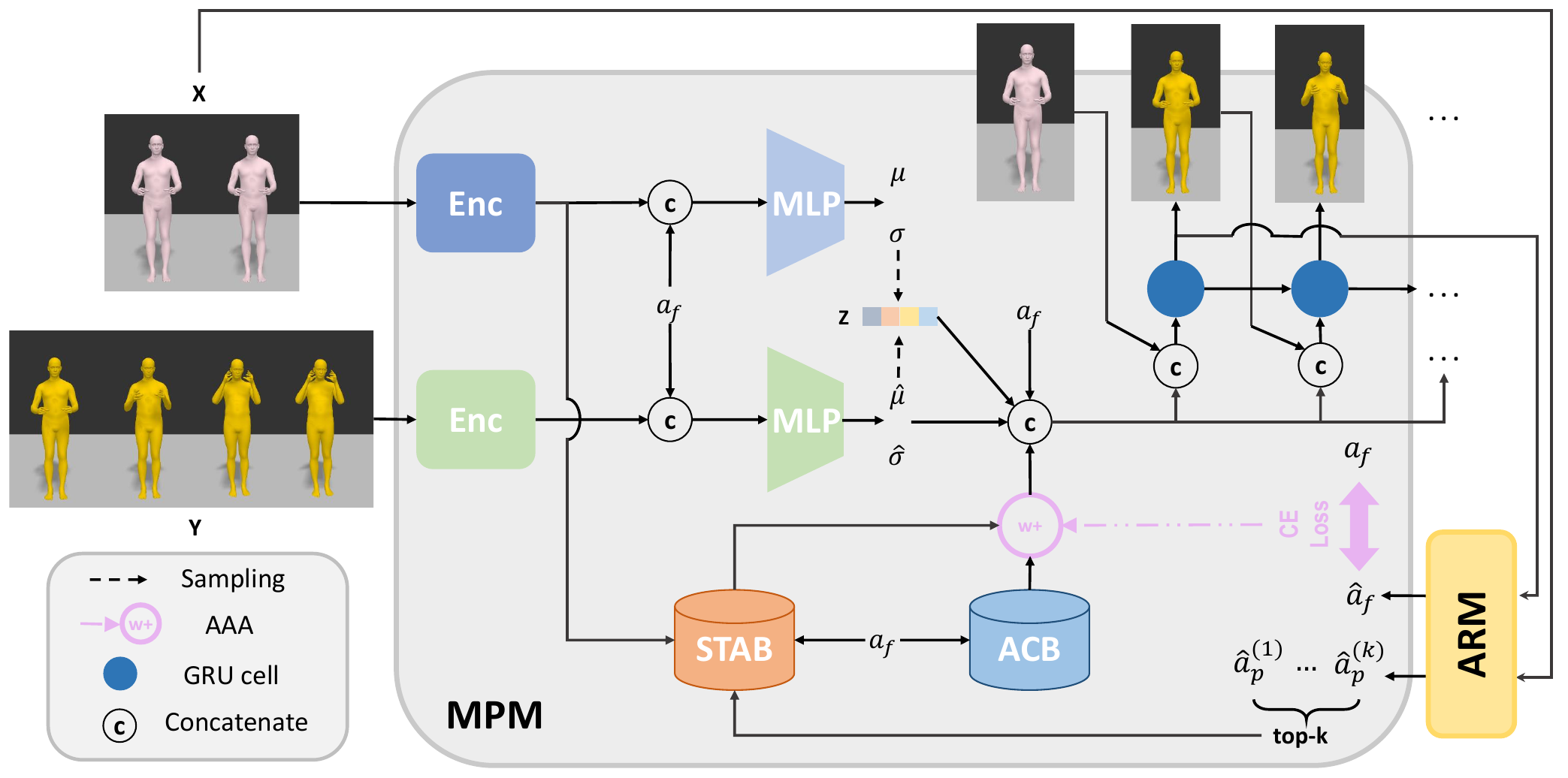}
   \vspace{-1.2mm}
   \caption{Overview of our model. The ``Enc'' and ``MLP'' represent encoders and multi-layer perceptrons that are the same as \cite{mao2022weakly}. ARM is our action recognition module and CELoss denotes the cross entropy loss. STAB is the soft-transition action bank and ACB is the action characteristic bank. $\mathbf{X}$ is the observed motion sequences and $\mathbf{Y}$ is the future ground-truth motions.}
   \vspace{-1.2mm}
   \label{fig:overview}
\end{figure*}
\section{Related Work}
\label{sec:formatting}
\noindent\textbf{Human Motion Prediction.}
Most prior work in human motion prediction focuses on forecasting motion within very short time frames, primarily differing in temporal encoding strategies, with approaches employing either recurrent structures \cite{fragkiadaki2015recurrent,gopalakrishnan2019neural,gui2018adversarial,jain2016structural,martinez2017human,pavllo2018quaternet,wang2019imitation} or feed-forward models \cite{aksan2019structured,butepage2017deep,li2018convolutional,mao2020history,mao2019learning}. Hu \etal \cite{DBLP:journals/pami/HuZLZ17} propose a three-step iterative optimization algorithm to find the optimal solution with a guaranteed convergence. \cite{Tang2023Predicting} uses recurrent attention mechanisms to transit temporal information. \cite{cai2020learning} constructs an action-agnostic bank representing the motion of each human joint. Tang \etal \cite{tang2023temporal} propose the prior compensation factor to mitigate the forgetting problem of prior knowledge during training. However, these methods generally lack semantic alignment, as they do not aim to generate motions linked to specific action labels from the same historical sequence. While some early works \cite{fragkiadaki2015recurrent,gopalakrishnan2019neural,gui2018adversarial,jain2016structural,martinez2017human} incorporated action information to support future predictions consistent with past motions, this information was limited and did not account for semantic variability. Additionally, previous methods \cite{bouazizi2022motionmixer,martinez2017human,li2018convolutional,tang2018long,corona2020context,hernandez2019human,mao2019learning,sofianos2021space,zhong2022spatio} adopted deterministic prediction strategies, which restricted motion diversity.

Recent approaches emphasize generating diverse motion outputs. Mainstream methods now often employ deep generative models, such as VAEs and GANs \cite{kingma2013auto,goodfellow2020generative}, to model the distribution of future motions. These models, typically built on encoder-decoder architectures \cite{salzmann2022motron,lucas2022posegpt,blattmann2021behavior,dang2022diverse,xu2022diverse}, leverage multi-loss constraints and masked completion \cite{aliakbarian2020stochastic,gupta2018social,lee2017desire,chen2023humanmac} to enhance diversity and realism. For example, DLow \cite{yuan2020dlow} focuses on sampling diverse future predictions from pre-trained generative models. While these approaches achieve notable diversity and plausibility, they do not incorporate explicit semantic categories. Our work follows the design of the task \cite{mao2022weakly}, to predict future motions across various action categories while ensuring both motion diversity and semantic alignment.

%-------------------------------------------------------------------------
\noindent\textbf{Human Motion Synthesis.}
The objective of human motion synthesis is to generate realistic human motion sequences without relying on prior observations. Early research primarily focused on modeling simple cyclic motions, employing methods such as principal component analysis \cite{ormoneit2005representing} or Gaussian process latent variable models \cite{urtasun2007modeling}. More recently, deep learning-based approaches \cite{ahuja2019language2pose,guo2020action2motion,lee2019dancing,li2020learning,li2021learn,lin2018human,petrovich2021action,shlizerman2018audio} have enabled the synthesis of more complex and nuanced motions, with some methods addressing inherent uncertainty in future motions \cite{bhattacharyya2018accurate,yan2018mt} to enhance generative accuracy. Diffusion models have widely adopted for 3D human motion modeling, leveraging U-Nets \cite{chen2023executing,dabral2023mofusion}, Transformers \cite{zhang2022motiondiffuse,wang2023fg}, and custom architectures \cite{zhang2023remodiffuse,alexanderson2023listen} to achieve high fidelity and flexibility in motion generation. Rajendran \etal \cite{rajendran2024review} provide a detailed review on synergizing the Metaverse and AI-driven synthetic data.

Beyond unconditional motion generation, conditioning on text descriptions allows for more refined control over generated outcomes \cite{ahuja2019language2pose,ren2023diffusion,wang2023fg,wei2023understanding,zhang2022motiondiffuse,zhao2023modiff,diller2024cg}. Additionally, some studies incorporate auxiliary signals such as audio or music \cite{lee2019dancing,li2020learning,li2021learn,shlizerman2018audio} to guide the generation. Other approaches use action labels as conditions, such as frame-level VAE models based on action labels \cite{guo2020action2motion} and sequence-level VAE models integrated with Transformers \cite{petrovich2021action}. While these methods are similar to \cite{mao2022weakly} in leveraging action labels, they primarily focus on generating specific actions from these labels, whereas \cite{mao2022weakly} centers on action prediction, utilizing historical motion information for future predictions.

%-------------------------------------------------------------------------
\noindent\textbf{Action-driven Stochastic Human Motion Prediction.}
Action-conditioned video generation aims to control video content using specific action labels, enabling precise alignment of target actions with visual output. For instance, \cite{yang2018pose} aims to predict future 2D human poses from a single input image and a target action label. In contrast to the setup in \cite{mao2022weakly}, \cite{yang2018pose} which is limited to a single input image, cannot capture action transitions. The approach in \cite{mao2022weakly} focuses on predicting multiple plausible future motions based on a sequence of action labels and brief motion history. By combining sequences from different actions and introducing weak supervision, \cite{mao2022weakly} designs a VAE model conditioned on observed motion and action sequences, enabling the generation of diverse future motions with varying lengths.
% \newpage
\section{Our Approach}%%%%%GRU less
%Let us now introduce our method to action-driven stochastic human motion prediction. 
In this paper, we represent 3D human motion using the SMPL model \cite{loper2023smpl}, which intends to parameterize a 3D human mesh with shape and pose parameters. Since we focus on human motion rather than human identity, we follow the implementations in \cite{mao2022weakly} and mainly predict the pose parameters of SMPL. Specifically, given an action label represented by an one-hot vector $\mathbf{a}$, and a sequence of $N$ past human poses represented by $\mathbf{X}=[\mathbf{x}_1, \mathbf{x}_2, \cdots, \mathbf{x}_N]\in \mathbb{R}^{K\times N}$, where $\mathbf{x}_i\in\mathbb{R}^{K}$ is the pose in the $i$-th frame, our goal is to predict a future motion $\hat{\mathbf{Y}}=[\hat{\mathbf{y}}_1, \hat{\mathbf{y}}_2, \cdots, \hat{\mathbf{y}}_T]\in \mathbb{R}^{K\times T}$, where $\hat{\mathbf{y}}_i\in\mathbb{R}^{K}$ represents predicting motion of $i$-th frame.

\subsection{Overview}
As Figure \ref{fig:overview} shows, our model contains two main parts, the Action Recognition Module (ARM) and Motion Prediction Module (MPM). ARM aims to classify the motion sequences, which is based on the gated recurrent unit (GRU). MPM bases on conditional VAEs (CVAEs) \cite{kingma2013auto} structure, aiming to predict future motion sequences. To utilize the action transition features and extract action characteristics better, we propose two memory banks, the soft-transition action bank (STAB) and action characteristic bank (ACB). Furthermore, we design the adaptive attention adjustment (AAA) strategy, aiming to fuse features from two banks effectively. As for training, we first train the ARM to infer the action of input observed motion sequence. Then, the MPM is optimized based on the classification results of ARM. This training strategy encourages the MPM to learn semantic information of different actions more easily.

\subsection{Action Recognition Module}
The target of ARM is to classify the action label for the observed motion sequence and let us denote the output of ARM as $\hat{\mathbf{a}}$. Since the input sequence can have different lengths, we use a one-layer GRU and an output multi-layer perceptron (MLP) with softmax as the backbone network to generate classifying results. This design allows us to classify sequences of any length, making it suitable for the modules we will introduce later.

\subsection{Motion Prediction Module}
We construct our Motion Prediction Module (MPM) in the framework of CVAE. Specifically, we design two memory banks, the soft-transition action bank (STAB) and action characteristic bank (ACB), to utilize the action transition features and extract action characteristics better. STAB stores the feature of action transitions, while ACB stores the motion features of a certain category action. Furthermore, we propose the adaptive attention adjustment (AAA) strategy to adaptively aggregate the contextual knowledge retrieved from two banks, thereby making the model's prediction results more accurate and reasonable.
\subsubsection{CVAE Structure}
Our model builds on CVAEs to predict action-driven future motions, as shown in Figure \ref{fig:overview}. The encoder and decoder structures follow \cite{mao2022weakly}. The goal is to model the conditional distribution $p(\mathbf{Y}|\mathbf{X}, \mathbf{a})$, where $q_{\phi}(\mathbf{z}|\mathbf{Y, X, a})$ models the posterior and $p_{\theta}(\mathbf{Y}|\mathbf{z, X, a})$ reconstructs future motions $\mathbf{Y}$ from latent variable $\mathbf{z}$. The evidence lower bound (ELBO) is:
\begin{equation}
    \begin{split}
        \log p(\mathbf{Y}|\mathbf{X, a}) \geq &\mathbb{E}_{q_{\phi}(\mathbf{z}|\mathbf{Y, X, a})}[\log p_{\theta}(\mathbf{Y}|\mathbf{z, X, a})] - \\
        & KL(q_{\phi}(\mathbf{z}|\mathbf{Y, X, a}) || p_{\psi}(\mathbf{z}|\mathbf{X, a})),
    \end{split}
\end{equation}
where $p_{\psi}(\mathbf{z}|\mathbf{X, a})$ is the prior, modeled by a network with parameters $\psi$. The $KL(\cdot || \cdot)$ term measures divergence.

Maximizing $\log p(\mathbf{Y}|\mathbf{X, a})$ involves maximizing the ELBO. The KL term is computed as:
\begin{equation}
    \begin{split}
        \mathcal{L}_{KL} &= KL(\mathcal{N}(\mu, diag(\sigma^{2})) || \mathcal{N}(\hat{\mu}, diag(\hat{\sigma}^{2}))) \\
        &= \frac{1}{2}\sum_{i=1}^{D}(\log \frac{\hat{\sigma_{i}}^2}{\sigma_{i}^2} + \frac{\sigma_{i}^2 + {(\mu_{i} - \hat{\mu_{i}})}^2}{\hat{\sigma_{i}}^2} - 1),
    \end{split}
\end{equation}
where $\mathcal{N}(\mu, diag(\sigma^{2}))$ and $\mathcal{N}(\hat{\mu}, diag(\hat{\sigma}^{2}))$ are the posterior and prior distributions. The parameters of these distributions are derived from the encoder $q_{\phi}$ and prior $p_{\psi}$, with $D$ being the latent dimension.

During training, $\mathbf{z}$ is sampled from the posterior using the reparameterization trick \cite{kingma2013auto}, \ie, $\mathbf{z}=\epsilon \odot \sigma + \mu$, where $\epsilon\sim\mathcal{N}(\mathbf{0, I})$. Given $\mathbf{z}$, past motion $\mathbf{X}$, and action label $\mathbf{a}$, the decoder $p_{\theta}$ reconstructs future motion. The reconstruction term in ELBO is:
\begin{equation}
    \mathcal{L}_{rec} = \frac{1}{T}\sum_{i=1}^{T}||\hat{\mathbf{y}_{i}} - \mathbf{y}_i||^{2}_{2},
\end{equation}
where $\hat{\mathbf{Y}} = [\hat{\mathbf{y}_{1}}, \hat{\mathbf{y}_{2}}, \cdots, \hat{\mathbf{y}_{T}}]$ represents future poses generated by the decoder.

Since $\mathbf{z}$ relies on ground truth $\mathbf{Y}$ during training, it is sampled from the prior $p_{\psi}(\mathbf{z}|\mathbf{X, a})$ in the inference stage.

\subsubsection{Soft-transition Action Bank}
\begin{figure*}[htbp]
  \centering
  % \fbox{\rule{0pt}{2in} \rule{0.95\linewidth}{0pt}}
   \includegraphics[width=0.75\linewidth,height=0.35\linewidth]{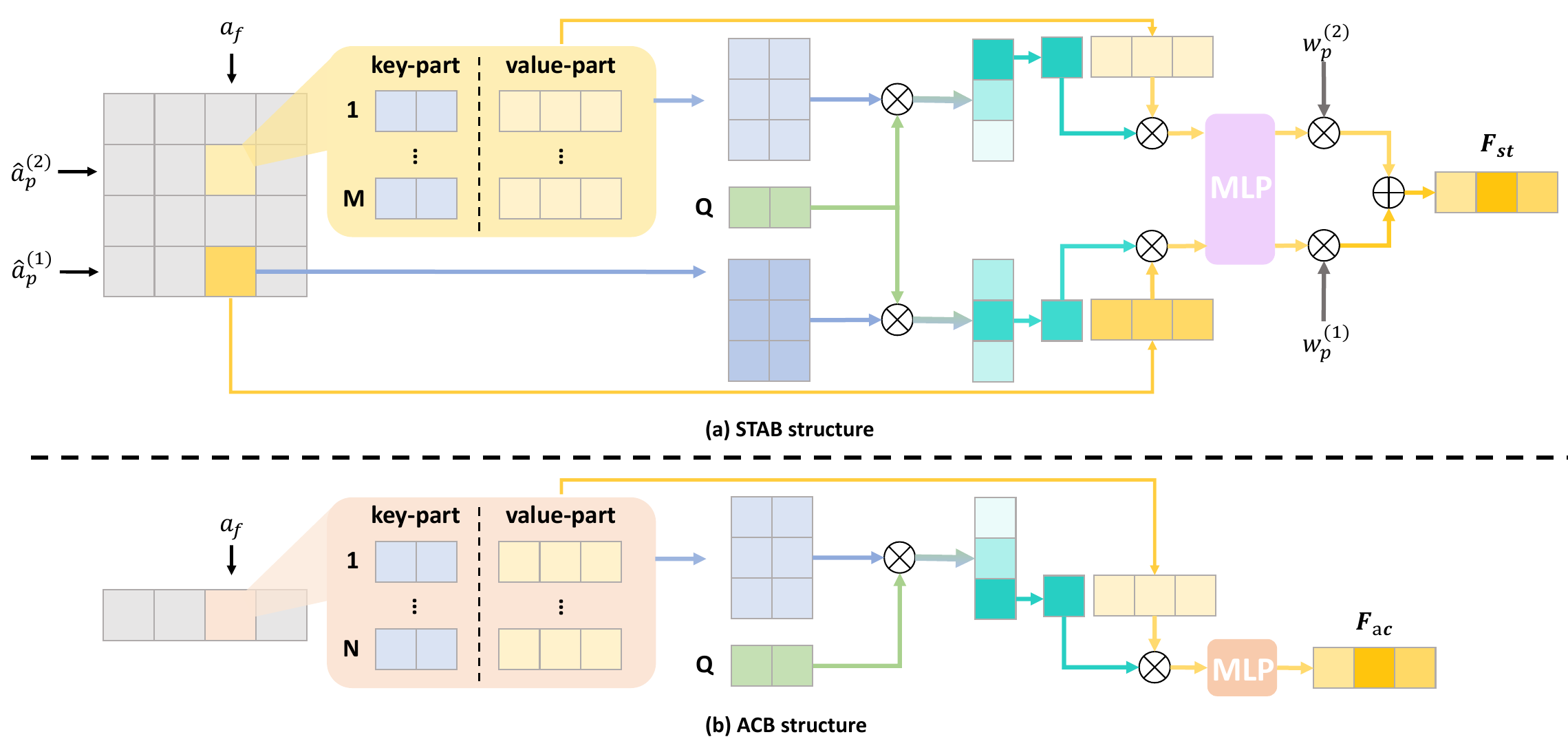}
   \vspace{-1.2mm}
   \caption{STAB and ACB structure. In (a), $\hat{a}_p$ and $a_f$ represent the past action label and future action label. Using these two labels, key-parts are retrieved from the bank, and similarity is computed between encoder-derived query features Q and the key-parts. The key-part with the highest similarity is selected, and its corresponding value-part feature is retrieved. A similarity-driven incentive strategy encourages the model to focus on bank features that closely match the query features. Similarly, (b) follows the same process.}
   \vspace{-2.2mm}
   \label{fig:stmbmfb}
\end{figure*}

To achieve smoother, more stable, and reasonable transitions between different generated actions, we design a bank based on action transitions. By querying this bank, the model can focus on the detailed changes in transitions between different actions. The details of our soft-transition action bank (STAB), as illustrated in Figure \ref{fig:stmbmfb}(a), are introduced below.

This bank is a set contains several elements that can be written as $\mathbf{S}_{st}=\{\mathbf{S}^{\hat{a}_p,a_f}\}$, where $\hat{a}_p$ and $a_f$ represent the past action label and future action label, respectively. Note that the past action label is given by the ARM. Each element is a set of $M$ tuples, that is $\mathbf{S}^{\hat{a}_p,a_f}=\{\mathbf{S}^{\hat{a}_p,a_f}_{1}, \mathbf{S}^{\hat{a}_p,a_f}_{2}, \cdots, \mathbf{S}^{\hat{a}_p,a_f}_{M}\}$. Each tuple is a pair of feature vectors, denoted as $\mathbf{S}^{\hat{a}_p,a_f}_{j} = (\mathbf{K}^{\hat{a}_p,a_f}_{j}, \mathbf{V}^{\hat{a}_p,a_f}_{j})$, in which $\mathbf{K}^{\hat{a}_p,a_f}_{j}$ and $\mathbf{V}^{\hat{a}_p,a_f}_{j}$ represent the key-part and the value-part, respectively. The searching processes of the bank can be divided into two steps. During the first search, we use the past and future action labels as indexes to find the corresponding $\mathbf{S}^{\hat{a}_p,a_f}$ within the bank. The second step involves calculating the similarity between the query features $\mathbf{Q}$ obtained from the encoder and the key-part $\mathbf{K}^{\hat{a}_p,a_f}_{i}$ of each tuple. Then, we choose the largest one and get its corresponding value-part feature $\mathbf{V}^{\hat{a}_p,a_f}_{i}$. We further design a similarity incentive strategy to encourage the model to learn more about the bank features that are more similar to the query features. We use the calculated similarity to weight the queried bank features, thereby giving more attention to more relevant features, which can be written as:
\begin{equation}
\label{equa:STB1}
    \begin{split}
        % &\mathbf{S}^{\hat{a}_p,a_f} = \mathbf{S}_{st}[\hat{a}_p][a_f] \\
        &\mathbf{Sim}_i = \mathbf{Q} \times \mathbf{K}^{\hat{a}_p,a_f}_i, \quad i\in\{1,2,\cdots,M\} \\
        &\mathbf{MaxS}, MaxI = \max(\mathbf{Sim}_1, \mathbf{Sim}_2, \cdots,\mathbf{Sim}_M)\\ 
        &\mathbf{F}_{st} = MLP(\mathbf{MaxS} \times \mathbf{V}^{\hat{a}_p,a_f}_{MaxI}),
    \end{split}
\end{equation}
where $\mathbf{Sim}_i$ is the similarity between the query features and the key-part of $i$-th tuple. $\mathbf{MaxS}$ and $MaxI$ represent the maximum similarity value and the corresponding index. $MLP$ denotes the multi-layer perceptron. The $\mathbf{F}_{st}$ is the output of soft-transition action bank.

However, due to the similarities between different actions, past motions can be classified into different categories. We need to consider different possible actions to make full use of the conditional information. Therefore, we innovatively adjusted the first searching step to a soft searching approach. Specifically, we use the top-$k$ recognized results from ARM instead of using the largest one and use these action categories as past action labels to query the STAB. Finally, the query results are weighted based on the softmax values of these top-$k$ categories, weighting the features obtained from different branch queries. The top-$k$ recognized results are denoted as $\mathbf{\hat{a}}_p = \{\hat{a}_p^{(1)}, \hat{a}_p^{(2)}, \cdots, \hat{a}_p^{(k)}\}$, and the corresponding weights are represented by $\mathbf{w}_p = \{w_p^{(1)}, w_p^{(2)}, \cdots, w_p^{(k)}\}$. Therefore, Equation (\ref{equa:STB1}) can be rewritten as:
\begin{equation}
\label{equa:STB2}
    \begin{split}
        % &\mathbf{\hat{a}}_p = \{\hat{a}_p^{(1)}, \hat{a}_p^{(2)}, \cdots, \hat{a}_p^{(k)}\} \\
        % &\mathbf{w}_p = \{w_p^{(1)}, w_p^{(2)}, \cdots, w_p^{(k)}\} \\
        % &\mathbf{S}^{\hat{a}_p^{(j)},a_f} = \mathbf{S}_{st}[\hat{a}_p^{(j)}][a_f] \\
        &\mathbf{Sim}^{(j)}_i = \mathbf{Q} \times \mathbf{K}^{\hat{a}_p^{(j)},a_f}_i, \quad i\in\{1,2,\cdots,M\} \\
        &\mathbf{MaxS}^{(j)}, MaxI^{(j)} = \max(\mathbf{Sim}^{(j)}_1,  \cdots,\mathbf{Sim}^{(j)}_M)\\ 
        &\mathbf{F}^{(j)}_{st} = MLP(\mathbf{MaxS}^{(j)} \times \mathbf{V}^{\hat{a}_p^{(j)},a_f}_{MaxI^{(j)}}) \\
        &\mathbf{F}_{st} = \sum_{j=1}^{k}w_p^{(j)}*\mathbf{F}^{(j)}_{st},
    \end{split}
\end{equation}
where $\hat{a}_p^{(j)}$ is the $j$-th largest category label of classifier output, and $w_p^{(j)}$ is the corresponding value.
\subsubsection{Action Characteristic Bank}
To extract the motion semantic feature of different actions, we design an action characteristic bank (ACB). Figure \ref{fig:stmbmfb}(b) illustrates the details of our action characteristic bank (ACB), which are introduced below.

Similar to STAB, ACB also contains some elements, which can be denoted as $\mathbf{S}_{ac}=\{\mathbf{S}^{a_f}\}$. $a_f$ is the future action label. Each element is a set of $N$ tuples, that is $\mathbf{S}^{a_f}=\{\mathbf{S}^{a_f}_{1}, \mathbf{S}^{a_f}_{2}, \cdots, \mathbf{S}^{a_f}_{N}\}$. Each tuple is a pair of feature vectors, denoted as $\mathbf{S}^{a_f}_{j} = (\mathbf{K}^{a_f}_{j}, \mathbf{V}^{a_f}_{j})$, in which $\mathbf{K}^{a_f}_{j}$ and $\mathbf{V}^{a_f}_{j}$ represent the key-part and the value-part, respectively. The retrieval step involves calculating the similarity between the query features $\mathbf{Q}$ obtained from the encoder and the key-part $\mathbf{K}^{a_f}_{i}$ of each tuple. Then, the largest one will be chosen. The similarity incentive strategy is then applied to its corresponding value-part feature $\mathbf{V}^{a_f}_{i}$ to generate the output feature. To be specific, the pipeline above can be written as:
\begin{equation}
\label{equa:MFB1}
    \begin{split}
        % &\mathbf{S}^{a_f} = \mathbf{S}_{ac}[a_f] \\
        &\mathbf{Sim}_i = \mathbf{Q} \times \mathbf{K}^{a_f}_i, \quad i\in\{1,2,\cdots,N\} \\
        &\mathbf{MaxS}, MaxI = \max(\mathbf{Sim}_1, \mathbf{Sim}_2, \cdots,\mathbf{Sim}_N)\\ 
        &\mathbf{F}_{ac} = MLP(\mathbf{MaxS} \times \mathbf{V}^{a_f}_{MaxI}),
    \end{split}
\end{equation}
where $\mathbf{Sim}_i$ is the similarity between the query features and the key-part of $i$-th tuple. $\mathbf{MaxS}$ and $MaxI$ represent the maximum similarity value and the corresponding index. $MLP$ denotes the multi-layer perceptron. The $\mathbf{F}_{ac}$ is the output of action characteristic bank.
\subsubsection{Adaptive Attention Adjustment}
Obviously, the features required for predicting motion at different time points are different. For example, in the initial period of prediction, we need to refer more to the features of the conditional frames to ensure smoother action transitions in the generated results. In the later period of prediction, we need to focus on the features of future actions to generate more accurate action sequences. Furthermore, due to differences in the rate and duration of changes in different actions, the changing process of each action is different. Therefore, we design an Adaptive Attention Adjustment (AAA) strategy to fuse the features $\mathbf{F}_{st}$ and $\mathbf{F}_{ac}$ effectively. Specifically, we use a parameter $\alpha$ to weight the outputs of STAB and ACB, which can be expressed as:
\begin{equation}
    \mathbf{F} = \frac{\alpha}{1+\alpha} * \mathbf{F}_{st} +  \frac{1}{1+\alpha} * \mathbf{F}_{ac},
\end{equation}
where $\mathbf{F}$ is the fused feature. The $\alpha$ is calculated from the cross entropy loss between the ARM classification result of the predicted frame and the ground truth label, which denotes $CELoss$. However, due to the difficulty in correctly judging actions when there are too few observation frames, we set a time threshold $\tau$ and only start changing $\alpha$ when the predicted time step is greater than the threshold. Furthermore, to avoid causing drastic fluctuations in the changes of $\alpha$ that could affect the training process, we used the running-mean method to modify the value of $\alpha$. The value of $\alpha$ can be expressed as:
\begin{equation}
\alpha =
\begin{cases}
     1, & t < \tau \\
     \gamma * \alpha + (1 - \gamma) * (CELoss - \alpha), & t \geq \tau
\end{cases}
\end{equation}
where $\gamma$ is a hyperparameter and $t$ is the predicted time step. 

Then we concatenate $\mathbf{F}$ with encoder output and input it into the decoder to generate the current frame's predicted result.

\section{Experiments}
\label{Exp}
\subsection{Datasets}
We conduct extensive experiments on the GRAB\cite{brahmbhatt2019contactdb,taheri2020grab}, NTU RGB-D \cite{liu2019ntu, shahroudy2016ntu}, BABEL \cite{punnakkal2021babel}, and HumanAct12 \cite{guo2020action2motion} datasets. We use the data preprocessed in \cite{mao2022weakly} to test the effectiveness of our approach, where the sequence is provided with a single action label, except for the BABEL set \cite{punnakkal2021babel}.

\noindent\textbf{GRAB} \cite{brahmbhatt2019contactdb,taheri2020grab} contains data of 10 agents interacting with 51 different objects, which results in a total of 29 different actions. Following \cite{mao2022weakly}, we mainly test our method using the data corresponding to actions of Pass, Lift, Inspect, and Drink, which have a large number of action samples (more than 1400 with motion length from 100 to 501 frames). We follow the cross-subject experimental setting and use data of 8 subjects (S1-S6, S9, S10) for training, and the remaining 2 subjects (S7, S8) for testing. The dataset is expanded by downsampling the sequences with a frame ratio of 15-30.

\noindent\textbf{NTU RGB-D} \cite{liu2019ntu, shahroudy2016ntu} is dataset collected for RGB-D action recognition and prediction. Following \cite{guo2020action2motion}, a subset containing 13 actions of, with noisy SMPL parameters estimated by VIBE \cite{kocabas2020vibe}. Following \cite{mao2022weakly}, we split the dataset into training and testing by subjects and train the model to predict future motion sequences with 10 past frames.

\noindent\textbf{BABEL} \cite{punnakkal2021babel} is a subset of the AMASS dataset \cite{mahmood2019amass}. We use the preprocessed dataset in \cite{mao2022weakly}, containing 9643 training samples and 3477 testing samples with 20 actions. The length of motion sequences ranges from 30 to 300. All samples have been downsampled to 30 Hz.

\noindent\textbf{HumanAct12} \cite{guo2020action2motion} consists of 12 subjects performing 12 actions. We use the sequences performed by 2 subjects (P11, P12) for testing, and the remaining 10 subjects (P1-P10) for training. The frame number of each sequence ranges from 35 to 290. The motion sequences with less than 35 frames are not included in the experiments. Overall, our model is trained using 727 sequences and tested on 197 testing sequences. %Our model is trained to observe 10 frames to predict the future.
\begin{table}[t]
  \centering
  \renewcommand{\arraystretch}{1.2}
    \setlength{\tabcolsep}{0.5mm}
    % \scriptsize
    % \footnotesize
    \tiny
    \begin{tabular}{|c|c|ccccc|}
    \toprule
     \textbf{Data}& \textbf{Method} & \textbf{Acc}$\uparrow$ &$\mathbf{FID_{tr}}\downarrow$ & $\mathbf{Fid_{te}}\downarrow$ & $\mathbf{Div_{w}}\uparrow$ & \textbf{Div}$\uparrow$ \\
    \midrule
    \multirow{4}[2]{*}{\begin{sideways}GRAB\end{sideways}} & Act2Mot & $70.6^{\pm1.3}$ & $80.22^{\pm6.64}$ & $47.81^{\pm1.09}$ & $0.50^{\pm0.00}$ & $0.76^{\pm0.01}$ \\
          & DLow  & $67.6^{\pm0.7}$ & $127.49^{\pm6.90}$ & $\mathbf{22.71^{\pm2.79}}$ & $0.74^{\pm0.01}$ & $0.92^{\pm0.01}$ \\
          & ACTOR  & $83.0^{\pm0.3}$ & $62.68^{\pm1.26}$ & $114.85^{\pm3.46}$ & $1.06^{\pm0.00}$ & $1.04^{\pm0.00}$ \\
          & WAT   & $92.6^{\pm0.6}$ & $44.59^{\pm1.39}$ & $38.03^{\pm1.49}$ & $1.10^{\pm0.01}$ & $1.37^{\pm0.01}$ \\
          & ours  & $\mathbf{95.23^{\pm0.32}}$ & $\mathbf{43.39^{\pm1.00}}$ & $34.18^{\pm1.55}$ & $\mathbf{1.14^{\pm0.04}}$ & $\mathbf{1.43^{\pm0.04}}$ \\
    \midrule
    \midrule
    \multirow{4}[2]{*}{\begin{sideways}NTU\end{sideways}} & Act2Mot & $66.3^{\pm0.2}$ & $144.98^{\pm2.44}$ & $113.61^{\pm0.84}$ & $0.75^{\pm0.01}$ & $1.19^{\pm0.01}$ \\
          & DLow  & $70.6^{\pm0.2}$ & $151.11^{\pm1.25}$ & $157.54^{\pm1.62}$ & $0.97^{\pm0.00}$ & $1.21^{\pm0.00}$ \\
          & ACTOR  & $66.3^{\pm0.1}$ & $355.69^{\pm5.74}$ & $193.58^{\pm2.91}$ & $\mathbf{1.84^{\pm0.00}}$ & $2.07^{\pm0.00}$ \\
          & WAT   & $76.0^{\pm0.2}$ & $72.18^{\pm0.93}$ & $111.01^{\pm1.28}$ & $1.25^{\pm0.00}$ & $\mathbf{2.20^{\pm0.00}}$ \\
          & ours  & $\mathbf{80.50^{\pm0.52}}$ & $\mathbf{65.11^{\pm3.27}}$ & $\mathbf{101.07^{\pm3.47}}$ & $1.24^{\pm0.00}$ & $2.19^{\pm0.01}$ \\
    \midrule
    \midrule
    \multirow{4}[2]{*}{\begin{sideways}BABEL\end{sideways}} & Act2Mot & $14.8^{\pm0.2}$ & $42.02^{\pm0.40}$ & $37.41^{\pm0.47}$ & $0.79^{\pm0.01}$ & $1.10^{\pm0.01}$ \\
          & DLow  & $12.7^{\pm0.2}$ & $27.99^{\pm0.45}$ & $24.18^{\pm0.59}$ & $0.65^{\pm0.00}$ & $0.90^{\pm0.00}$ \\
          & ACTOR  & $40.9^{\pm0.2}$ & $29.34^{\pm0.10}$ & $30.31^{\pm0.16}$ & $\mathbf{2.94^{\pm0.00}}$ & $\mathbf{2.71^{\pm0.00}}$ \\
          & WAT   & $49.6^{\pm0.4}$ & $22.54^{\pm0.27}$ & $22.39^{\pm0.36}$ & $1.35^{\pm0.00}$ & $1.74^{\pm0.00}$ \\
          & ours  & $\mathbf{55.37^{\pm0.38}}$ & $\mathbf{20.35^{\pm0.19}}$ & $\mathbf{20.26^{\pm0.29}}$ & $1.35^{\pm0.00}$ & $1.72^{\pm0.00}$ \\
    \midrule
    \midrule
    \multirow{4}[1]{*}{\begin{sideways}HumanAct12\end{sideways}} & Act2Mot & $24.5^{\pm0.1}$ & $245.35^{\pm7.13}$ & $298.06^{\pm10.80}$ & $0.31^{\pm0.00}$ & $0.60^{\pm0.01}$ \\
          & DLow  & $22.7^{\pm0.2}$ & $254.72^{\pm8.48}$ & $143.71^{\pm3.07}$ & $0.35^{\pm0.00}$ & $0.53^{\pm0.00}$ \\
          & ACTOR  & $44.4^{\pm0.2}$ & $248.81^{\pm3.77}$ & $381.56^{\pm6.66}$ & $\mathbf{0.84^{\pm0.00}}$ & $0.95^{\pm0.00}$ \\
          & WAT   & $59.0^{\pm0.1}$ & $129.95^{\pm0.39}$ & $164.38^{\pm2.27}$ & $0.74^{\pm0.00}$ & $0.96^{\pm0.00}$ \\
          & ours  & $\mathbf{61.57^{\pm0.32}}$ & $\mathbf{112.85^{\pm3.80}}$ & $\mathbf{137.28^{\pm2.68}}$ & $0.76^{\pm0.00}$ & $\mathbf{1.02^{\pm0.01}}$ \\
    \bottomrule
    \end{tabular}%
    \caption{Quantitative results. %We report the action recognition accuracy (Acc), the FID to training data ($FID_{tr}$) and to the testing split ($FID_{te}$), and the diversity before (Div) and after DTW ($Div_{w}$). 
    Results of Act2Mot, DLow, ACTOR and WAT are from \cite{mao2022weakly}.}
    \vspace{-4.2mm}
  \label{tab:mainexp}%
\end{table}%
\subsection{Evaluation Metrics}
We follow the evaluation protocol of \cite{mao2022weakly} and employ the following metrics to evaluate our method:
\begin{itemize}
    \item To measure the distribution similarity between the generated sequences and ground-truth motions, we adopt the Frechet Inception Distance (FID) \cite{heusel2017gans}:
    \begin{equation}
    \begin{split}
        FID &= ||\mu_{gen} - \mu_{gt}||^2 \\
            &+ Tr(\Sigma_{gen} + \Sigma_{gt} - 2(\Sigma_{gen}\Sigma_{gt})^{\frac{1}{2}}),
    \end{split} 
    \end{equation}
    where $\mu_{.} \in \mathbb{R}^{d}$ and $\Sigma_{.} \in \mathbb{R}^{d\times d}$ are the mean and covariance matrices of perception features from the pre-trained action recognition model in \cite{mao2022weakly}. $d$ denotes the feature dimension, and $Tr(\cdot)$ represents the trace operator.

    \item The action recognition accuracy of the generated motions is reported using the same pre-trained action recognition model, which evaluates the motion realism of generated sequences.

    \item To evaluate per-action diversity, we measure the pairwise distance between multiple future motions generated from the same historical motion and action label. For a set of future motions $\{\hat{\mathbf{Y}}^{i}\}_{i=1}^{G}$ predicted by our model, diversity is calculated as:
    \begin{equation}
        Div = \frac{2}{G(G-1)}\sum_{i=1}^{G}\sum_{j=i+1}^{G}\frac{1}{T_{max}}\sum_{v=1}^{T_{max}}||\hat{\mathbf{y}}^{i}_{v} - \hat{\mathbf{y}}^{j}_{v}||_{2},
    \end{equation}
    where $T_{max}$ is the maximum number of predicted frames, and $\hat{\mathbf{y}}^{i}_{v}$ is the $v$-th frame of motion $\hat{\mathbf{Y}}^{i}$.

    Following \cite{mao2022weakly}, per-action diversity is computed after Dynamic Time Warping (DTW) \cite{zhang2019predicting}, which aligns motions temporally as $\hat{\mathbf{Y}}^{i}, \hat{\mathbf{Y}}^{j} = DTW(\hat{\mathbf{Y}}^{i}, \hat{\mathbf{Y}}^{j})$, where aligned motions $\hat{\mathbf{Y}}^{i}, \hat{\mathbf{Y}}^{j} \in \mathbb{R}^{K \times T_{i,j}}$ have the same number of frames $(T_{i,j})$. The diversity is then calculated as:
    \begin{equation}
        Div_{w} = \frac{2}{G(G-1)}\sum_{i=1}^{G}\sum_{j=i+1}^{G}\frac{1}{T_{i,j}}\sum_{v=1}^{T_{i,j}}||\hat{\mathbf{y}}^{i}_{v} - \hat{\mathbf{y}}^{j}_{v}||_{2}.
    \end{equation}
\end{itemize}
\begin{figure*}[t]
  \centering
  % \fbox{\rule{0pt}{2in} \rule{0.95\linewidth}{0pt}}
   \includegraphics[width=0.75\linewidth]{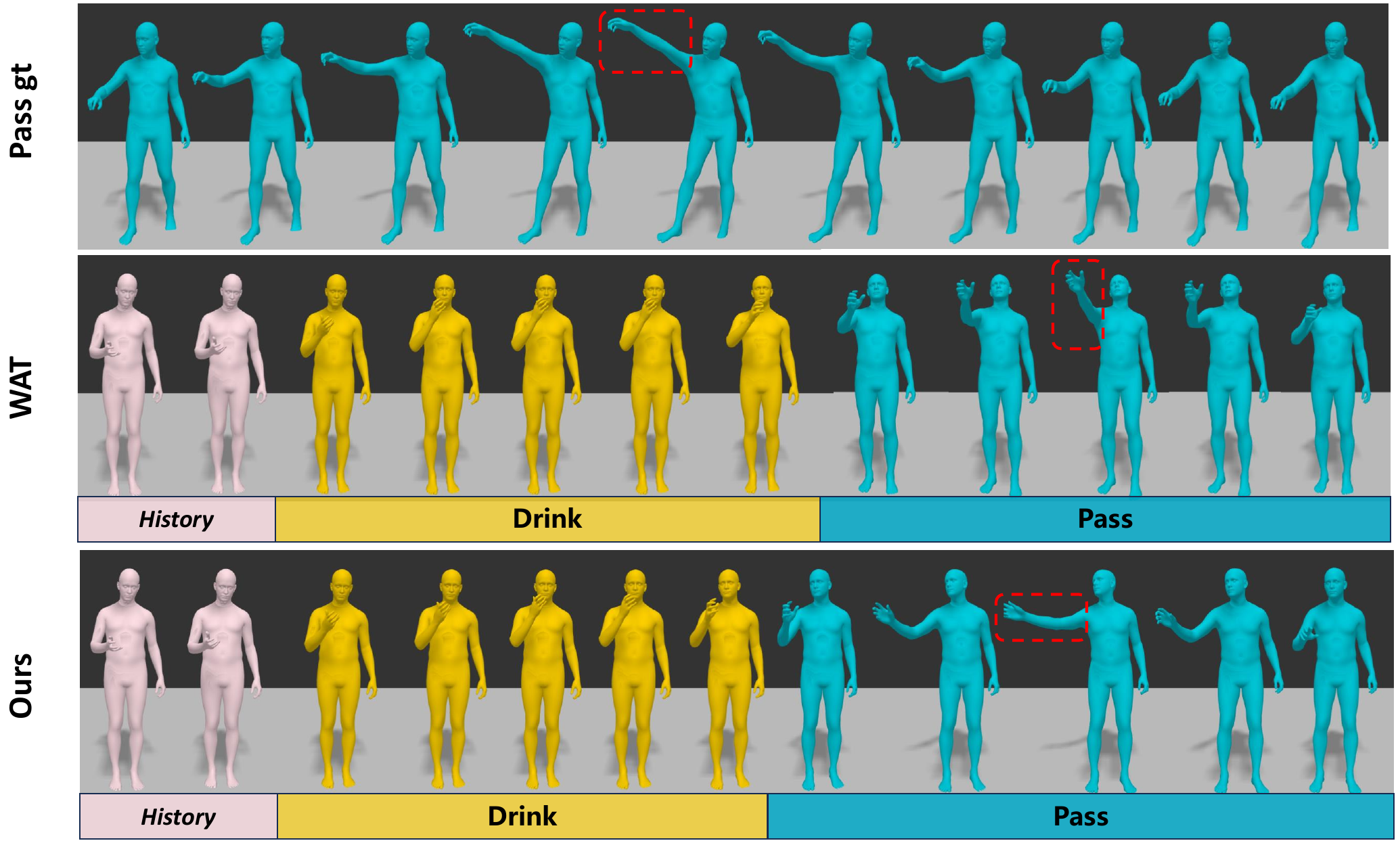}
   \vspace{-2.2mm}
   \caption{A visualization comparison between our model, WAT, and ground truth on the GRAB dataset. This example illustrates the generation of a future ``Drink'' action sequence (yellow) and the transition to a ``Pass'' action sequence (blue), conditioned on past ``Drink'' action sequences (pink). WAT produces a future ``pass'' action with a forward arm movement, whereas our model yields actions that more closely align with the ground truth, achieving higher accuracy in motion detail.}
   \vspace{-3.2mm}
   \label{fig:visual}
\end{figure*}
\subsection{Implementation Details}
%\noindent \textbf{Training of ARM}
We first train the ARM for 500 epochs using ADAM \cite{diederik2014adam} optimizer with an initial learning rate of 0.002. The learning rate is fixed for the first 50 epochs and gradually decreases thereafter. We use cross entropy (CE) loss to train ARM. To be specific, we add the CE loss to every frame with a time step greater than $\tau$. When ARM is well trained, we freeze the parameters of ARM and begin training MPM. We also train the MPM with 500 epochs using the ADAM optimizer. The learning rate is initialized with 0.002 and gradually decreases after 100 epochs. We add the CE loss to the original loss function in \cite{mao2022weakly}. Specifically, the generated motion sequences are classified by our ARM. Then, we compute the CE loss between ARM output and ground-truth labels.
%\noindent \textbf{Training of MPM}
\subsection{Experimental Results}
% Table generated by Excel2LaTeX from sheet 'Sheet1'
We compare our approach with previous SOTA methods, including Act2Mot \cite{guo2020action2motion}, DLow \cite{yuan2020dlow}, ACTOR \cite{petrovich2021action} and WAT \cite{mao2022weakly}. Act2Mot introduces frame-wise motion VAE with GRUs. DLow focuses on generating diverse motions, while WAT proposes a transition learning model for action-driven stochastic motion prediction.

Table \ref{tab:mainexp} presents the experimental results across multiple datasets. Our method achieves the best performance on all four datasets. Specifically, on GRAB and HumanAct12, our approach significantly outperforms WAT across all metrics. For NTU and Babel, our method surpasses WAT in both Acc and FID metrics, with a minor gap of less than 0.01 in the Div metric on NTU. This small difference falls within acceptable error margins, demonstrating the robustness of our approach.
The improvements are attributed to the proposed STAB and ACB modules. STAB captures rich action transition information and extracts diverse transition features through soft search. ACB retrieves the inherent motion characteristics of each action. By adaptively fusing these components, our method effectively enhances both prediction accuracy and diversity.
\subsection{Visualization}
\begin{figure*}[t]
  \centering
  % \fbox{\rule{0pt}{2in} \rule{0.95\linewidth}{0pt}}
\includegraphics[width=0.75\linewidth]{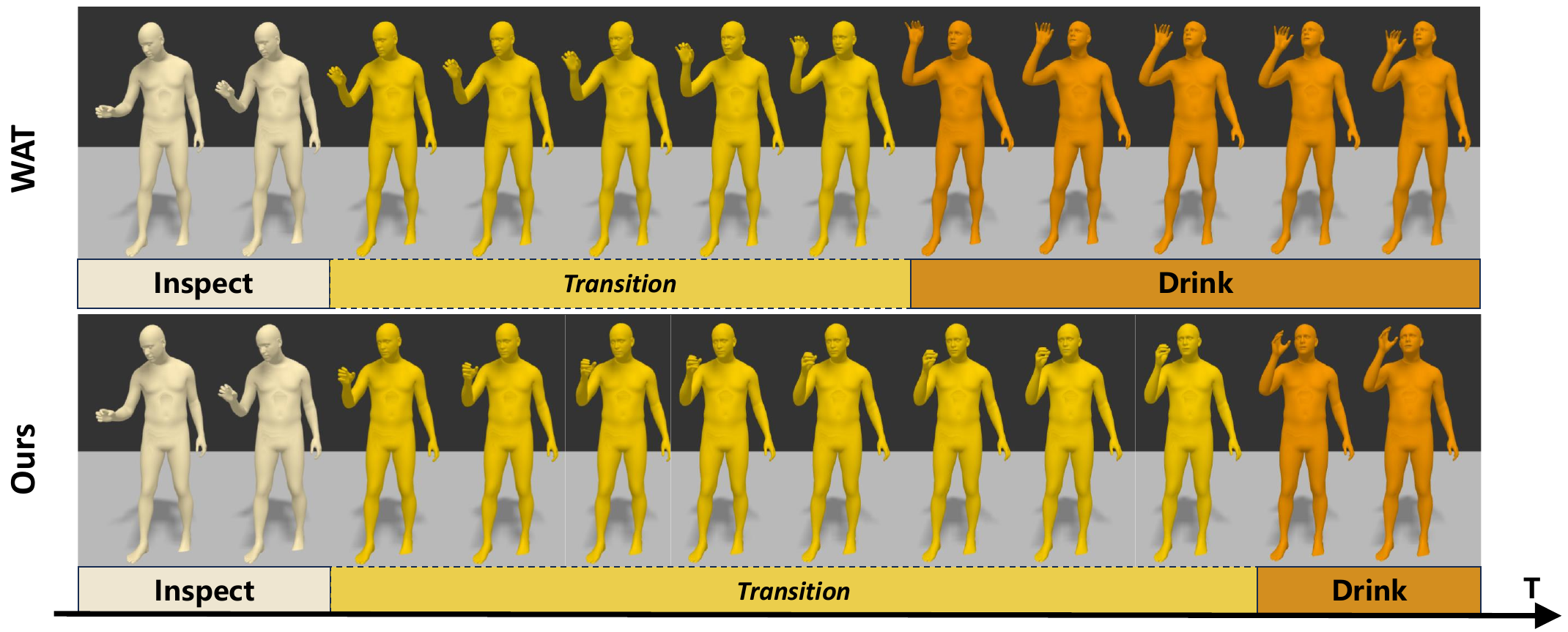}
\vspace{-2.2mm}
   \caption{This visualization compares action transitions generated by WAT and our model, focusing on the sequence from ``Inspect'' (off-white) to ``Drink'' (orange). Transitions are uniformly sampled at a consistent frame interval. WAT demonstrates a noticeably rigid transition. In contrast, our model generates additional transition actions, resulting in a smoother and more natural motion.}
   \vspace{-3.2mm}
   \label{fig:visual2}
\end{figure*}
Here, we present some qualitative results, illustrating the action quality generated by our method. Figure \ref{fig:visual} provides a sample from the GRAB dataset. As shown, this example sequence transitions from the historical ``Drink'' action to the predicted ``Pass'' action. Notably, WAT \cite{mao2022weakly} generates a forward arm movement in the ``Pass'' action, while our model produces a more accurate sequence that closely aligns with the ground truth. This demonstrates that our ACB module enhances the accuracy of action generation. In Figure \ref{fig:visual2}, we illustrate an ``inspect'' to ``drink'' action transition. WAT \cite{mao2022weakly} exhibits abrupt and rigid movements with large spans. In contrast, through the STAB module, our model generates additional sequences between actions, resulting in smoother and more natural transitions.  
\subsection{Ablation Studies}
We conduct ablation experiments on the GRAB dataset to further verify the effectiveness of our proposed framework. 
\begin{table}[t]
  \centering
  \renewcommand{\arraystretch}{1.5}
    \setlength{\tabcolsep}{0.5mm}
    \scriptsize
    % \footnotesize
    % \tiny
    \begin{tabular}{|c|ccccc|}
    \toprule
    \multicolumn{1}{|c|}{\textbf{Method}} & \textbf{Acc}$\uparrow$ & $\mathbf{FID_{tr}}\downarrow$ & $\mathbf{Fid_{te}\downarrow}$ & $\mathbf{Div_{w}}\uparrow$ & \textbf{Div}$\uparrow$ \\
    \midrule
    w/o AAA & $91.93^{\pm0.49}$ & $44.00^{\pm2.01}$ & $35.50^{\pm1.20}$ & $1.10^{\pm0.04}$ & $1.40^{\pm0.04}$ \\
    w/o STAB & $92.18^{\pm0.60}$ & $43.97^{\pm2.75}$ & $35.71^{\pm2.06}$ & $1.11^{\pm0.03}$ & $1.40^{\pm0.03}$ \\
    w/o ACB & $93.45^{\pm0.50}$ & $43.61^{\pm1.80}$ & $34.58^{\pm2.20}$ & $1.13^{\pm0.03}$ & $1.43^{\pm0.02}$ \\
    w/o RM & $90.84^{\pm0.90}$ & $48.99^{\pm3.66}$ & $38.70^{\pm2.49}$ & $1.11^{\pm0.04}$ & $1.41^{\pm0.05}$ \\
    \hline
    ours  & $\mathbf{95.23^{\pm0.32}}$ & $\mathbf{43.39^{\pm1.00}}$ & $\mathbf{34.18^{\pm1.55}}$ & $\mathbf{1.14^{\pm0.04}}$ & $\mathbf{1.43^{\pm0.04}}$ \\
    \bottomrule
    \end{tabular}%
    \vspace{-1.2mm}
    \caption{Results of the ablation study on the proposed module. %``w/o AAA'' represents model without the adaptive attention adjustment strategy. ``w/o STAB'' is the model without the soft-transition action bank. ``w/o ACB'' denotes model without the action characteristic bank. ``w/o RM'' represents changing $\alpha$ without using the running mean method.%
    }
    \vspace{-3.2mm}
  \label{tab:abla1}%
\end{table}%
% Table generated by Excel2LaTeX from sheet 'Sheet1'
\subsubsection{Ablation of proposed module}
% Table generated by Excel2LaTeX from sheet 'Sheet1'
Table \ref{tab:abla1} shows the results of ablation study, highlighting the importance of our AAA strategy, STAB, ACB, the dynamic adjustment parameter $\alpha$ and the running-mean operation to change $\alpha$. Without the AAA strategy, replacing it with a mean value of query features leads to performance degradation across all metrics, especially in accuracy. This occurs because the model overemphasizes action transition information during later prediction stages, increasing the gap between observed motion and future actions, which ultimately reduces generation accuracy. This demonstrates the effectiveness of AAA in dynamically adjusting the model's attention to different features. Similarly, removing STAB causes a significant drop in accuracy, as information stored in the bank is critical for guiding smoother transitions in the early stages of prediction, resulting in more reasonable and accurate outputs. Excluding ACB also reduces performance, though it has minimal impact on diversity, indicating that STAB contributes to maintaining diversity in the generated results. Finally, as ``w/o RM'' shows, modifying $\alpha$ without running mean method introduces noise and instability during early training, hindering model convergence and significantly lowering performance.
\vspace{-4.2mm}
\subsubsection{The impact of $k$}
We evaluate the impact of different values of $k$, as shown in Table \ref{tab:abla2}. Four $k$ values are tested. When $k=1$, the soft-transition method is not applied, resulting in inaccurate action transitions and degraded prediction performance. As $k$ increases, prediction accuracy improves initially but declines afterward, while diversity steadily increases. This trend is expected: smaller $k$ values lack sufficient variation to handle early similar actions, whereas larger $k$ values query excessive tokens, leading to overly similar features and less informative results. While larger $k$ introduces greater diversity by incorporating more variations, excessive diversity can negatively impact prediction quality.
\begin{table}[t]
  \centering
  \renewcommand{\arraystretch}{1.1}
    \setlength{\tabcolsep}{0.5mm}
    % \scriptsize
    \footnotesize
    % \tiny
    \begin{tabular}{|c|ccc|}
    \toprule
    \multicolumn{1}{|c|}{\textbf{Method}} & \textbf{Acc}$\uparrow$ & $\mathbf{Div_{w}}\uparrow$ & \textbf{Div}$\uparrow$ \\
    \midrule
    top-1 & $93.03^{\pm0.75}$ & $1.10^{\pm0.04}$ & $1.39^{\pm0.04}$ \\
    top-2 & $\mathbf{95.23^{\pm0.32}}$ & $1.14^{\pm0.04}$ & $1.43^{\pm0.04}$ \\
    top-3 & $93.74^{\pm0.59}$ & $1.21^{\pm0.03}$ & $1.50^{\pm0.04}$ \\
    top-4 & $92.57^{\pm0.54}$ & $\mathbf{1.22^{\pm0.03}}$ & $\mathbf{1.52^{\pm0.04}}$ \\
    \bottomrule
    \end{tabular}%
    \vspace{-2.2mm}
    \caption{Results of evaluating different $k$ in the STAB.}
    \vspace{-4.2mm}
  \label{tab:abla2}%
\end{table}%
\section{Conclusion}
\label{Conclu}
In this paper, we propose two novel memory banks to address challenges in action-driven stochastic human motion prediction. The STAB records transition features, with a soft searching approach encouraging focus on multiple possible action categories in observed motions. The ACB stores action characteristics, enhancing the quality of generated motions. Additionally, the AAA strategy improves feature fusion, further boosting model performance. Extensive experiments validate the effectiveness of our method, achieving SOTA results on four motion prediction datasets.\vspace{-3.2mm}
\paragraph{Limitations.} The proposed bank introduces a slight increase in memory and computation overhead. Using WAT as baseline, frames per second (FPS) decrease from 2.76 to 1.98 and GPU memory consumption increases from 2262 to 2837 MB. However, efficiency can be further enhanced by implementing parallel retrieval of bank elements.
\vspace{-1.2mm}
\section*{Acknowledgments}
\vspace{-1.2mm}
This work was partially supported by NSFC (U21A20471, 62476296), Guangdong Natural Science Funds Project (2023B1515040025, 2022B1515020009), Guangdong Provincial Science and Technology Program Project (2024A1111120017), and Guangdong Provincial Key Laboratory of Information Security Technology (2023B1212060026).
{
    \small
    \bibliographystyle{IEEEtran}
    \bibliography{main}
}

% WARNING: do not forget to delete the supplementary pages from your submission 
% \input{sec/X_suppl}

\end{document}